# TOWARDS A LARGE-SCALE FUSED AND LABELED DATASET OF HUMAN POSE WHILE INTERACTING WITH ROBOTS IN SHARED URBAN AREAS

**Elahe Sherafat**
Ph.D. Student
Laboratory of Innovations in Transportation (LiTrans)
Toronto Metropolitan University
Toronto, Canada

**Bilal Farooq (corresponding author)**
Associate Professor
Laboratory of Innovations in Transportation (LiTrans)
Toronto Metropolitan University
Toronto, Canada
Email: bilal.farooq@torontomu.ca

Word Count: 3823 words + 3 table(s) × 250 = 4573 words

## ABSTRACT

Over the last decade, Autonomous Delivery Robots (ADRs) have transformed conventional delivery methods, responding to the growing e-commerce demand. However, the readiness of ADRs to navigate safely among pedestrians in shared urban areas remains an open question. We contend that there are crucial research gaps in understanding their interactions with pedestrians in such environments. Human Pose Estimation is a vital stepping stone for various downstream applications, including pose prediction and socially aware robot path-planning. Yet, the absence of an enriched and pose-labeled dataset capturing human-robot interactions in shared urban areas hinders this objective. In this paper, we bridge this gap by repurposing, fusing, and labeling two datasets, MOT17 and NCLT, focused on pedestrian tracking and Simultaneous Localization and Mapping (SLAM), respectively. The resulting unique dataset represents thousands of real-world indoor and outdoor human-robot interaction scenarios. Leveraging YOLOv7, we obtained human pose visual and numeric outputs and provided ground truth poses using manual annotation. To overcome the distance bias present in the traditional MPJPE metric, this study introduces a novel human pose estimation error metric called Mean Scaled Joint Error (MSJE) by incorporating bounding box dimensions into it. Findings demonstrate that YOLOv7 effectively estimates human pose in both datasets. However, it exhibits weaker performance in specific scenarios, like indoor, crowded scenes with a focused light source, where both MPJPE and MSJE are recorded as 10.89 and 25.3, respectively. In contrast, YOLOv7 performs better in single-person estimation (NCLT seq 2) and outdoor scenarios (MOT17 seq1), achieving MSJE values of 5.29 and 3.38, respectively.

## INTRODUCTION

The shift in customer behavior towards online purchasing has led to a surge in demand for front-door delivery. This change is evident in the substantial increase in online purchases, rising from %5.6 of total retail sales in 2009 to %16 in 2019. With the onset of the Covid pandemic, online purchases escalated even further, reaching %33 in 2020. Notably, China has experienced remarkable growth in online shopping and on-demand food delivery, with annual increases of %30 to %50 (*1*).

As a result, Delivery Robots have emerged as a viable remedy for tackling the workforce shortage and rising delivery expenses. This development is not only reshaping the delivery industry but also influencing traffic patterns (*2*). These entities have made substantial investments in technologies aimed at lowering delivery costs and enhancing operational efficiency (*3*).

Autonomous Delivery Robots (ADRs), commonly referred to as mobile delivery robots and Personal Delivery Devices (especially in regulatory contexts), are vehicles that transport cargo and engage with individuals outside of them, such as cyclists and pedestrians. These vehicles are either fully or partially automated, eliminating the need for drivers or on-site delivery personnel. They are designed to operate on sidewalks, crosswalks, and road shoulders, primarily confined to a restricted delivery area, like a college campus.

Like autonomous vehicles (AVs), these vehicles are equipped with sensors such as cameras, LiDAR (Light Detection and Ranging), and radar to gather information from their surroundings. Additionally, they utilize global position devices and Inertial Measurement Units (IMU) to navigate their immediate environment.

Training Autonomous Delivery Robots to make their navigation in urban areas safe is highly challenging due to the diverse range of factors that come into play. Until these challenges are effectively addressed, widespread deployment of these robots in urban areas and shared spaces will remain limited to pilot programs.

Because Sidewalk Autonomous Delivery Robots (SADRs) travel on sidewalks and have the same right-of-way as pedestrians, they have been the subject of increasing regulation by local agencies in the U.S. (*3*). These challenges originated from robots dealing with infrastructure, interaction with objects, humans, and pets in shared urban areas, Figure 1, and their limitations at the operational level.

For the widespread implementation of Autonomous Delivery Robots in urban settings, it is crucial to prioritize their safe path planning while also considering social awareness. Human pose estimation and prediction serve as the fundamental building blocks for achieving this goal.

## HUMAN POSE ESTIMATION LITERATURE

Human pose estimation is defined as the process of assigning the positions of the joints in a human body, given an image, video, or a series of images capturing that individual (*5*). While some papers may use the terms "pose prediction" and "pose estimation" interchangeably, there is a significant distinction at the core of these two concepts. Predicting human pose necessitates a pose representation of the human body, which is accomplished through a pose estimation task. In other words, RGB images and videos (sequences of frames at various time steps) serve as input to a pose estimation network, and the network is responsible for labelling the input to reveal the human pose, represented as body shape or skeleton, for further human motion analysis.

Hence, a pose estimation algorithm receives an image as input and produces the corresponding pose. The resulting sequence of poses for an individual or a group of people is treated as time series data, which will then be fed into a pose prediction network to anticipate the poses of people in future time steps.

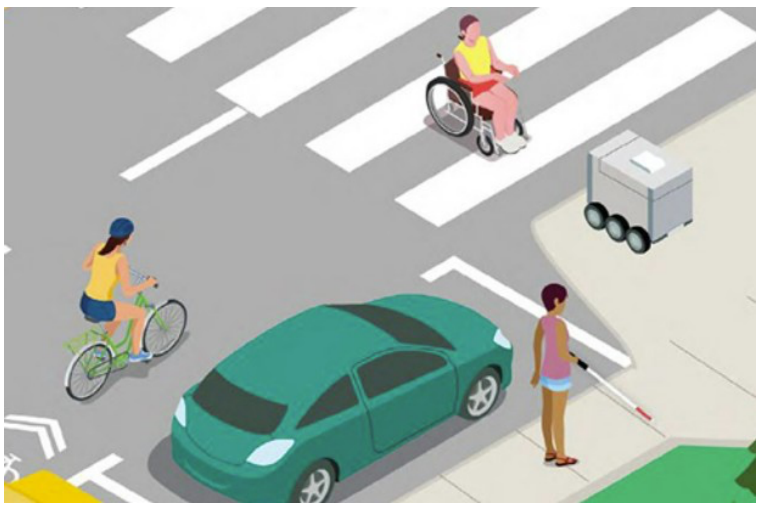

**FIGURE 1**: **A diagram depicting the interaction of SADR with pedestrians with various characteristics in shared urban areas, such as crosswalks, sidewalks, and bike lanes. (*4*))**

pose estimation finds diverse applications in transportation, such as pedestrian intention estimation (*6*), understanding pedestrian behavior (*7*), and its utilization in autonomous vehicles and robots (*8, 9*). However, the limited availability of enriched datasets has hindered extensive investigations into their application in autonomous delivery robots.

In contrast to *human trajectory estimation*, which has been studied in prior research, Kalatian and Farooq (*10*), the problem of *human pose estimation* is considerably more intricate. It encounters challenges related to a larger feature space, making it a more unsolved and complex task. Compared to the trajectory problem, which represents humans as a single trajectory point coordination, the pose estimation problem is much more informative by providing multiple coordination for each human joint.

As a result of concise and detailed body representation, robots could go beyond just collision avoidance and be enabled to infer humans' head orientation, intentions, and social interaction. They can distinguish individuals in wheelchairs, kids, and any pose-related inference. Having accurate pose data, robots can adhere to social rules; they would be enabled to detect conversations happening in their path and to choose routes that do not interrupt group conversations. That is known as the socially aware path planning of robots that would not be possible without human pose labeled datasets.

Combined learning from many datasets is an important research area since deep learning-based models perform best when fed large amounts of data with different themes and characteristics (*11*). In our context, although there are numerous labeled human pose datasets, none have the element of human-robot interaction. At the same time, we have service and delivery robots operating worldwide in close interaction with pedestrians. These robots could highly benefit from the dataset to improve their efficiency. To ensure they are not a hazard to humans and, at a higher level, their socially aware interaction, we must first analyze humans' behavior in the presence of robots. Pose estimation is a crucial task and the stepping stone that all behavioral models build upon it.

## DATASETS

In this study, we aimed to gather and fuse all the open-access datasets that capture human-robot interaction in shared environments , either directly or indirectly. To the best of our knowledge, there is no open-access data currently available that directly targets this problem. However, we could find some datasets that were initially collected for other purposes, e.g. Simultaneous Localization and Mapping (SLAM) Carlevaris-Bianco et al. (*12*) or pedestrian tracking Milan et al. (*13*) purposes. In this study, those datasets were repurposed and labeled using YOLOv7. Particularly, the University of Michigan North Campus Long-Term (NCLT) Vision and LIDAR Dataset Carlevaris-Bianco et al. (*12*) and the Multiple Object Tracking MOT17 benchmark, by Milan et al. (*13*) datasets are investigated, and the scenes of our interest collected and fused. The fused data is fed into the YOLOv7, and the visualized and numeric results are evaluated and human-annotated following a consistent protocol.

### NCLT dataset

The NCLT is a long-term and large-scale dataset collected by a Segway robotic platform 2a bi-weekly between January 2012 and April 2013 on the University of Michigan's North Campus. The dataset contains scenes that capture real-world scenes of pedestrians in the presence and with a robot's perspective in different indoor and outdoor environments.

This dataset is initially collected for Simultaneous Localization and Mapping (SLAM) application and aimed at human-infrastructure interaction while considering pedestrians, if any, as the dynamic obstacles. However, we could find subsets of sequences of our interest in different lighting conditions and directions of pedestrians facing the moving robot in finite and shared indoor and outdoor spaces.

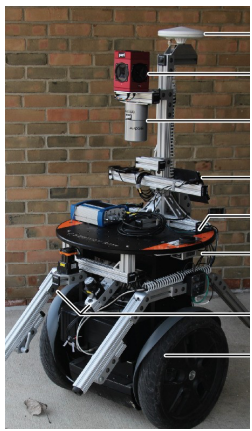 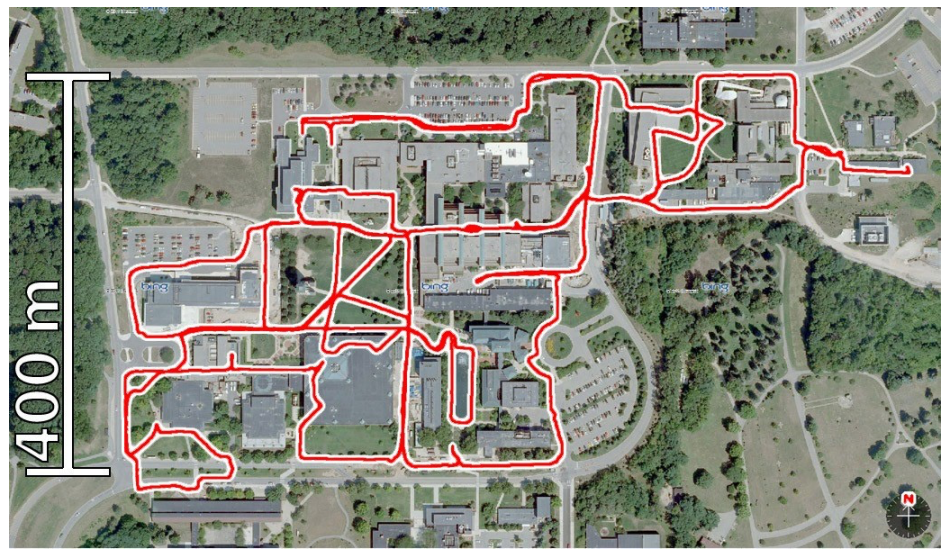

(a) The Segway robot equipped with data collection sensors, cameras, and GPS

(b) Robots trajectory passing from indoor and outdoor environments

**FIGURE 2**: **The NCLT data collection details. Carlevaris-Bianco et al. (*12*)**

Table 1 describes the only six available human-robot interaction sequences in the total 147.4 km of the robot's trajectory in the 27 sessions at the campus site. The human-robot interaction subset contains only 251 frames. However, they are enriched with different lighting conditions, indoor and outdoor environments, infrastructure geometry and crowd level, and direction of human dynamics.

**TABLE 1**: **Elaboration of the sequences within the human-robot subset of the NCLT dataset.**

| Sequence | Scenario | Frames | Environment |
|---|---|---|---|
| Seq 1 | Dark crowded aisle | 121 | Indoor |
| Seq 2 | Bright aisle with a single person | 24 | Indoor |
| Seq 3 | Dark, the approach of the robot to a crosswalk while two pedestrians are crossing in front of the robot | 26 | Outdoor |
| Seq 4 | Dark, the approach of the robot to an intersection, multi-direction of pedestrian | 28 | Outdoor |
| Seq 5 | Sever dark, single person | 28 | Outdoor |
| Seq 6 | Fixed location of the robot, crossing of pedestrians in front of the robot | 24 | Outdoor |

Figure 3 illustrates the scenes and scenarios captured by the Ladybug3 omnidirectional camera for each sequence within the human-robot subset of the NCLT dataset, as detailed in Table 1.

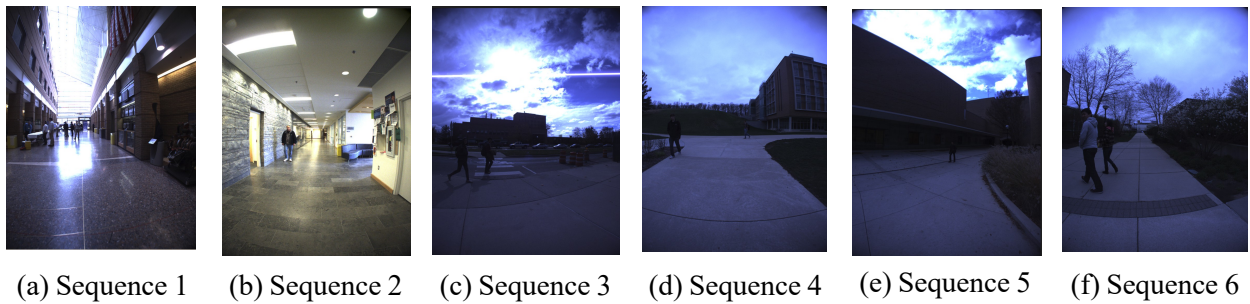

(a) Sequence 1     (b) Sequence 2     (c) Sequence 3     (d) Sequence 4     (e) Sequence 5     (f) Sequence 6

**FIGURE 3**: **Sample scenes of the human-robot interaction subset of the NCLT dataset**

**MOTChallenge: The Multiple Object Tracking Dataset**
The initial release of the benchmark occurred in 2014 and was subsequently updated to the MOT16 and MOT17 versions. These updates included the incorporation of new and existing single-camera object tracking datasets and are collected using various sources, including camera surveillance, a camera mounted on a vehicle on a road or a mobile platform in shared urban areas. The MOTchallenge particular emphasis is on multiple pedestrian tracking, as it remains one of the most extensively studied aspects in the field of object tracking (*13*).

The particular subset of MOT17 we are interested in is derived from a study focused on multi-person tracking using a mobile platform (*14*). The data was collected using a combination of stereo camera pair, LIDAR, and GPS, as illustrated in Figure 4. Using the stereo cameras suites the data set for further depth analysis (*15*), pedestrian detection (*16*) and tracking (*17*) purposes. However, no research has focused on human pose estimation or provided pose annotations for this dataset.

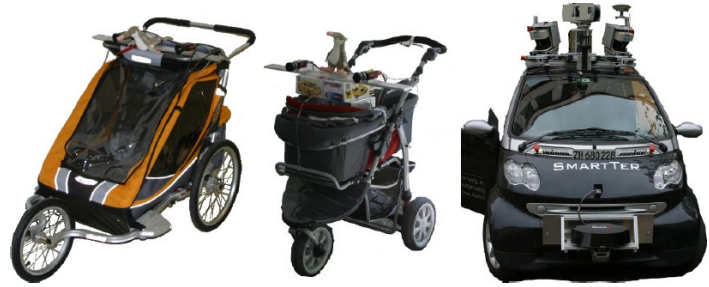

**FIGURE 4**: **The data collection devices equipped with stereo camera pair, Ess et al. (*14*)**

The data collected by stereo cameras mounted on strollers, as shown in the left and middle images of Figure 4, offers a perspective similar to that of delivery robots. These strollers are nearly identical in size and speed to delivery robots and serve as moving obstacles for pedestrians, akin to the behavior of delivery robots. Therefore, they would serve as a suitable representation of delivery robots, effectively capturing human behavior during interactions with these moving obstacles in confined shared urban spaces.

The subset consists of six sequences taking place in crowded urban scenes and one set in a sunny location. These scenes encompass diverse elements, such as kids, the elderly, Figure 5e, scooter riders, people walking in groups, and reflections of pedestrians in windows, Figure 5f. Additionally, the data is collected across various locations, including wide and narrow sidewalks, crosswalks, Figure 5a, and intersections. The subset details are elaborated in Table 3.

The diverse and dynamic scenes present in the data make it a suitable selection for analyzing human behavior during interactions with robots in urban environments. However, it is essential to be mindful of one limitation: the contrast in pedestrians' perception and reactions to a stroller, a common sight in urban areas, compared to an automated delivery robot, which is still relatively uncommon in many countries.

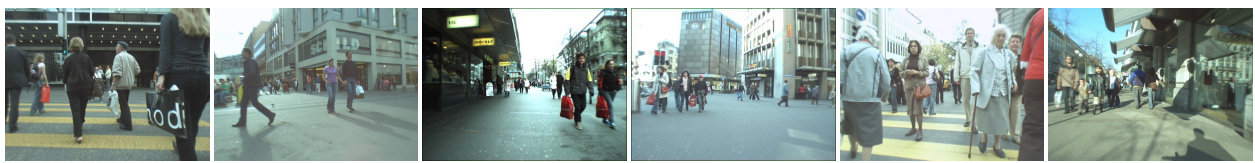

(a) Sequence 1    (b) Sequence 2    (c) Sequence 3    (d) Sequence 4    (e) Sequence 5    (f) Sequence 6

**FIGURE 5**: **Sample scenes of the human-robot interaction subset of the *MOT17* dataset**

## METHODOLOGY

In this study, we adhere to the framework depicted in Figure 6 with the objective of creating a robust human pose-labeled dataset that captures pedestrians' behavior and reactions in shared confined areas when encountering moving obstacles and automated robots. Initially, appropriate open-access datasets that could directly or indirectly contribute to the research were collected, fused, and pre-processed. Afterward, the 12 sequences are fed to the You Only Look Once (YOLOv7) network to generate desired outputs. The model provides bounding boxes surrounding pedestrians, joint locations of pedestrians, and RGB pose-labeled images.

**TABLE 2**: **Elaboration of the sequences within the human-robot subset of the MOT17 dataset.**

| Sequence | Scene's features | Frames | Environment |
|----------|------------------|--------|-------------|
| Seq 1 | Crossing a signalized intersection, entering a wide sidewalk | 220 | Intersection, sidewalk |
| Seq 2 | Highly crowded, Multi and bidirectional flow of pedestrians, robot's navigation through pedestrians approaching it, kids' presence, passing an unsignalized intersection | 1209 | Street car line, sidewalk, street |
| Seq 3 | Wild sidewalk with a bidirectional flow of pedestrians, pedestrians passing the sidewalk, and pedestrians entering and exiting stores. | 1000 | Sidewalk with a streetcar at the right |
| Seq 4 | Entering a wide pedestrian area with several entrances, a crosswalk at the left | 450 | Wide sidewalk area |
| Seq 5 | Approaching a crosswalk, waiting at the crosswalk with other pedestrians, crowded, seniors' existence, cyclists, a kid with a scooter, one and bidirectional and bidirectional flow of pedestrians in a narrow sidewalk. | 939 | Crosswalk, wide and narrow sidewalks |
| Sunny | Almost separated bidirectional flow of pedestrians in sunny weather, occlusion, walking in groups, kids' presence, people with stroller | 453 | Sidewalk |

A subset of four sequences, two from each dataset, were selected to provide the ground truth for the model evaluation. Comparing the joint locations generated by the model and the ground truth pose, the traditional MPJPE is calculated. Besides, by corporation the dimensions of the bounding boxes the novel proposed MSJE error is calculated to address the distance bias of the traditional one. Lastly, The HDP error is measured to access the human detection

**Pose Estimation**

You Only Look Once (YOLO) models are single-stage object detectors, and version 7 (YOLOv7) is the first version that is enabled to estimate human pose (*18*). Compared to other open source tools, especially, MediaPipe (*19*), MoveNet (*20*), and OpenPose (*21*), YOLOv7 performs the best when the input image has occlusion (*22*).

The YOLOv7 model estimates human pose by identifying 17 key joint locations of the human body and face. For this research, the model's numeric outputs are tailored to include the image number, the pedestrian's identification number, the x and y coordinates of the pedestrian's bounding box center, as well as the width and height of the bounding box. Furthermore, the outputs also encompass the x and y coordinates for each joint location of the pedestrian's body joints, all following a specific order. Besides, each point coordination is assigned with a confidence level.

Figure 7 illustrates the human pose output generated by the network for a sample from each subset. Figure 7a demonstrates the network's accuracy in handling challenging scenarios, including indoor and crowded scenes with poor lighting conditions, as well as humans positioned in groups at a considerable distance from the robot. In contrast, the right image, as depicted in Figure 7a, demonstrates the network's performance when objects are too close to the camera and lead to occlusion.

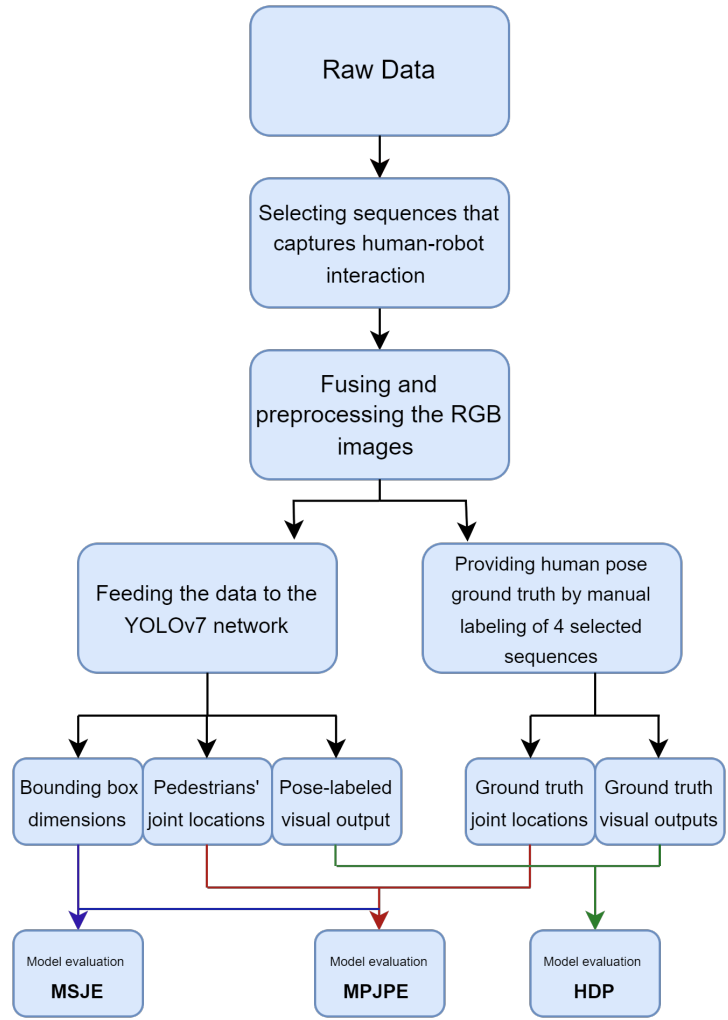

**FIGURE 6**: **The framework of the study.**

**Ground Truth Development**

In the context of human pose estimation, there will always be an error arising from the discrepancy between the correct (ground truth) joint coordinates and those estimated by the network. The magnitude of this error can vary depending on scene features such as density, pedestrian flow direction, image resolution, and lighting conditions. Furthermore, due to the error's unit being in pixels, it is significantly biased toward the distance from the camera.

Irrespective of the origin of errors and biases, it is essential to establish the ground truth or actual joint locations to calculate the discrepancy between the network's pose estimation and the true human pose. This can be achieved by manually labelling the RGB images, which serve as the same input as the network. Due to the absence of human pose annotations in the datasets, MOT17 and NCLT, experts from the Laboratory of Innovations in Transportation (LiTrans) manually annotated subsets of sequences following precise instructions to establish the ground truth.

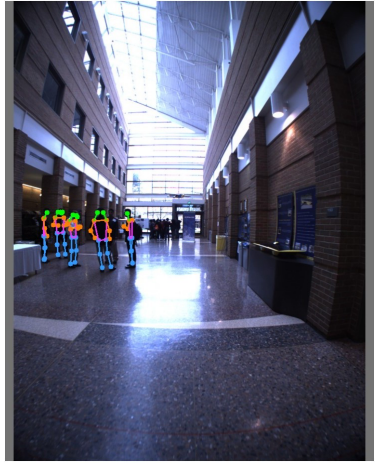

(a) The NCLT sample output

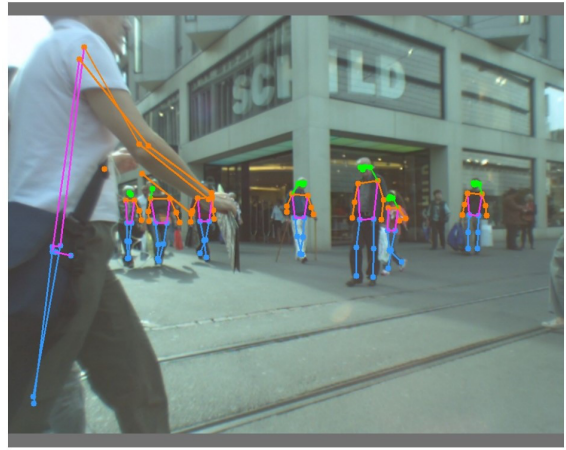

(b) The *Mot17* sample output

**FIGURE 7**: **The YOLOv7 visualized human pose output for the NCLT subset, in the left, and the _MOT17_ subset, in the right.**

The annotation protocol is precisely developed due to the various challenges associated with human pose labelling, including occlusion, partial visibility, crowded scenes, poor lighting conditions, reflections of pedestrians in mirrors and windows, cropped scenes, and the impact of clothing. These challenges introduce uncertainties that necessitate a comprehensive and unified annotation approach. Upon detecting a pedestrian, YOLOv7 assigns joint locations for all 17 human joints, even for joints that are occluded or do not exist in the scene due to the pedestrian entering or exiting the scene. However, the network assigns a lower confidence level to these joints.

When it comes to partially occluded joints, they are estimated in the manual labelling process, as the human brain is capable of estimating those joints. Nonetheless, joints that have yet to enter the scene or have exited the scene in previous time steps are excluded from the evaluation process.

By feeding the fused RGB images of the MOT17 and NCLT datasets into the YOLOv7 network, the model produced a total of 4271 and 254 numeric and visual outcomes for the MOT17 and NCLT datasets, respectively. Four subsets were specifically selected to evaluate YOLOv7's human pose estimation performance, and the ground truth numeric and visual pose labels were meticulously generated through manual labelling following unique instructions.

**Evaluation Metrics**

*Mean Per Joint Position Error (MPJPE)*

Developing an error metric to assess the model's performance is of utmost significance in the context of human pose estimation or prediction. In this study, we deployed three evaluation and error metrics. After reviewing the literature, it becomes evident that the Mean Per Joint Position Error (MPJPE) (*23*) has emerged as the most frequently employed evaluation metric (*24–27*).

The Mean Per Joint Position Error (MPJPE) computes the average Euclidean distance between the ground truth and estimated joint positions across all individuals in the sequence at each frame. This is depicted in Equation 1 (*23*). In the equation, $E_{MPJPE}(f, S)$ represents the error originating from frame $f$ and skeleton $S$, where $N_S$ denotes the number of joints for each human pose representation (in this case, 17). The function $m_{YOLO,S}^{(f)}(i)$ returns the coordinates of the $i$-th joint of skeleton $S$ at frame $f$, while $m_{gt,S}^{(f)}(i)$ represents the coordinates of that joint as a "ground truth" determined through the manual annotation process, as discussed in Section 5.2.

$$E_{MPJPE}(f, s) = \frac{1}{N_S} \sum_{i=1}^{N_S} ||m_{YOLO,S}^{(f)}(i) - m_{gt,S}^{(f)}(i)||^2 \tag{1}$$

### Mean Scaled Joint Error (MSJE)
Several research works have addressed inherent uncertainties in the fundamental MPJPE error metric, particularly occlusion and visibility issues. For example, Adeli et al. (*28*) introduced novel approaches such as the Visibility Ignored Metrics (VIM) and the visibility-aware metric (VAM) to tackle these concerns.

Nonetheless, the bias towards the distance from the camera still persists. In other words, two pedestrians with similar error values—one close to the robot and the other farther away—would have different implications. A one-pixel error in a specific joint location would have a higher impact on the pedestrian closer to the camera compared to the one farther away.

This study proposes a novel evaluation metric called Mean Scaled Joint Error (MSJE) to tackle this issue. MSJE scales the error by utilizing the width and height measurements of the bounding box surrounding the pedestrians, as shown in Equation 2.

$$E_{MSJE}(f, s) = \frac{1}{N_S} \sum_{i=1}^{N_S} \left[ \frac{||m_{YOLO,S}^{(f)}(i) - m_{gt,S}^{(f)}(i)||^2}{H_i \cdot W_i} \right] \tag{2}$$

### Human Detection Percentage (HDP)
The Human Detection Percentage (HDP) is calculated as the ratio of the number of detected pedestrians in a scene to the total number of pedestrians present in the scene, averaged across all frames. This error primarily focuses on evaluating the human detection capability of the network rather than pose estimation. However, it helps to assess the performance of YOLOv7 and determine its sensitivity to distance.

## RESULT AND DISCUSSION
This section presents the numeric and visualized results of the YOLOv7 human pose estimation. Figure 8 illustrates several visual outputs of the network. Notably, as shown in Figure 8a, the model exhibits accurate pose estimation in crowded areas with multiple directions of pedestrian flow. However, Figure 8b demonstrates the detection of pedestrian reflections in the middle of the scene on the right window, identified as pedestrians by the model. The network's limitation in handling human reflections on windows or mirrors can be mitigated by applying filters to the outputs.

Figure 8c demonstrates the network's proficiency in handling occlusion. As depicted, most of pedestrian 2's joint locations are obscured by pedestrian 1's right arm. Nonetheless, the model performs accurately in accurately estimating a significant portion of the joint locations. Figure 8d showcases the model's performance in an indoor aisle with poor lighting conditions. While the model effectively estimated the human pose for the group of individuals close to the robot, it failed to detect those standing at the end of the aisle and directly in front of the light source.

Figures 8e and 8f reveal the model's sensitivity to rotated images as input. When provided with 90-degree rotated images, its performance diminishes, as evidenced in Figure 8f, where the model detected the shadow of a pedestrian as an actual pedestrian. However, it illustrates the model's performance in dealing with severely dark environments, where the pose estimation is challenging even for humans.

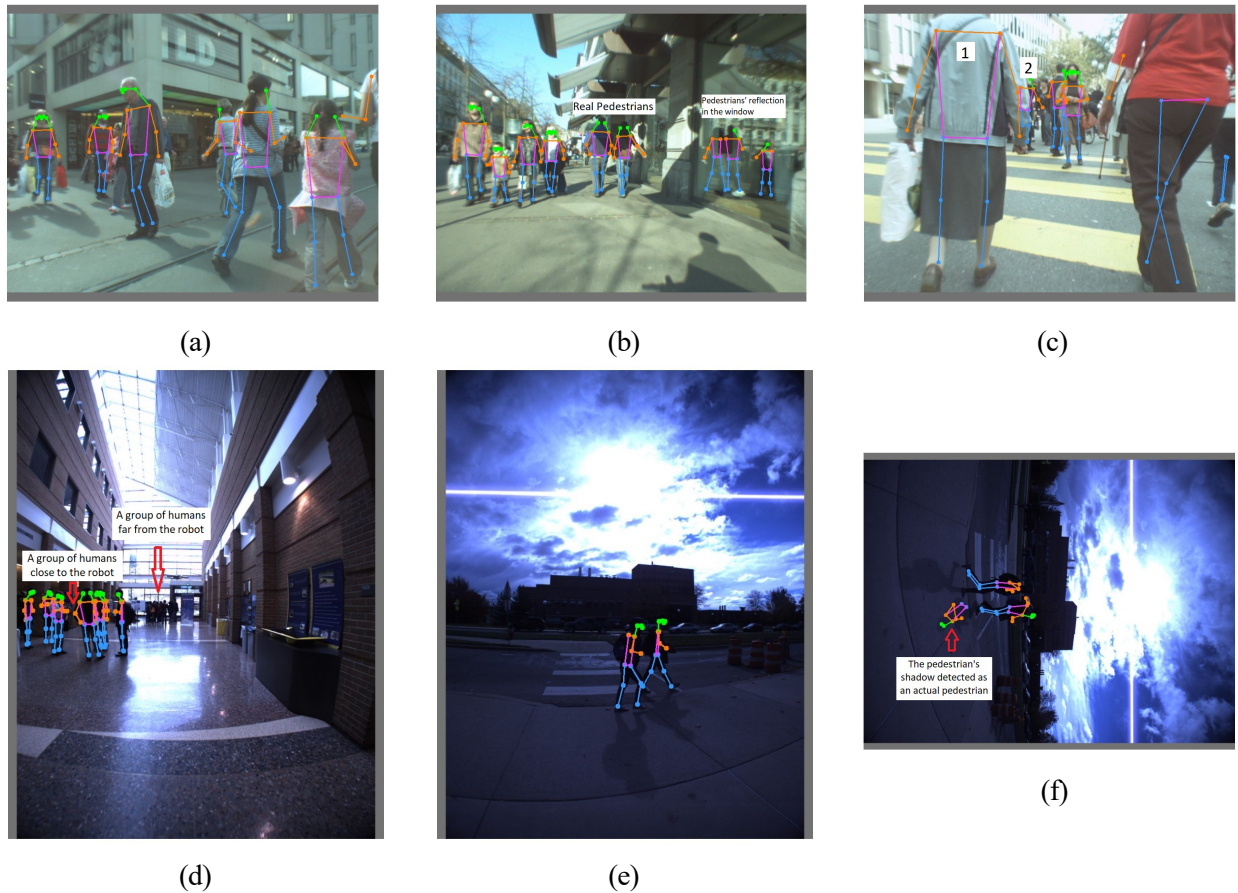

(a)                                         (b)                                         (c)

(d)                                         (e)                                         (f)

**FIGURE 8**: **Visualized output of YOLOv7 for human pose estimation in MOT17 (upper images) and NCLT (bottom images).**

Table 3 showcases the MPJPE, MSJE, and HDP metrics for sequences 1 and 2, which serve as subsets of the NCLT and MOT17 datasets, as explained in Table 1 and 3, respectively. In general, based on error metrics, the YOLOv7 model demonstrates reasonably good performance in human pose estimation for both datasets (NCLT and MOT17).

Figure 3a, 3b, 5a, and 5b illustrate the sample sequences NCLT seq 1, NCLT seq 2, MOT 17 seq 1, and MOT 17 seq 2, respectively. As depicted in Table 3, the network's performance, assessed by the naive MPJPE metric, showed the best for MOT17 seq2, with a loss of 5.79. Subsequently, NCLT seq2 exhibited a marginally higher error of 9.99. Notably, both datasets' sequence 1 yielded identical errors, each amounting to 10.89.

The intriguing finding is the significant change in both the error's magnitude and order when utilizing the MSJE metric, which scales the error based on the dimensions of the bounding box surrounding the pedestrians. Upon considering MSJE, Mot17 seq1 exhibited the smallest error of 3.38, followed by NCLT seq2 and MOT17 seq2, with errors of 5.29 and 8.8, respectively. Lastly, NCLT seq1 had the highest MSJE error, amounting to 25.3.

**TABLE 3**: **Evaluation of Human Pose Estimation for Selected Sequences - Error Metrics.**

| Sequence | Avg Height | SD Height | MPJPE | MSJE$\times 10^3$ | HDP |
|----------|-----------|-----------|-------|-------|-----|
| NCLT seq 1 | 128.44 | 32.53 | 10.89 | 25.3 | 58 |
| NCLT seq 2 | 190.28 | 77.81 | 9.99 | 5.29 | 100 |
| MOT17 seq 1 | 454.40 | 184.86 | 10.89 | 3.38 | 77 |
| MOT17 seq 2 | 223.35 | 127.21 | 5.79 | 8.8 | 27 |

According to both metrics, the model's performance in NCLT seq1 is the least favourable. The primary reason for this is the poor lighting condition resulting from the light source being located at the end of the indoor aisle, Figure 3a. Additionally, based on the average height, the detected humans in this sequence are positioned farther away from the camera. Hence the estimation error increase. This error exhibits its true magnitude compared to other sequences when scaled using the MSJE metric.

The lowest MSJE error is associated with MOT17seq1. This is due to the high-resolution images and the proximity to the camera for 77% of the pedestrians that the network detected. On the contrary, the remaining 23% of pedestrians were not recognized by the network because they were positioned at a much greater distance from the camera. This distance variation is a result of the robot stopping, with one group standing close behind the red traffic light, and the other group situated on the other side of the street, as depicted in Figure 5a. The highest standard deviation for the pedestrian height in this sequence confirms this variance in distances from the camera.

## CONCLUSION

This study introduced a novel fused and labeled pose dataset capturing humans interacting with moving obstacles and sidewalk robots in a confined shared urban area by repurposing MOT17 and NCLT datasets. The resulting dataset bridges the gaps and enables further research opportunities, including human pose prediction and socially aware path planning of robots that ensure their safe navigation in shared urban spaces.

The pose labels are generated using the YOLOv7 deep-learning vision network. The network's performance is evaluated in various scenarios through a meticulous manual labelling process following a concise protocol. Additionally, a new evaluation metric is proposed to tackle the distance bias in previous methods. This metric scales the error based on the dimensions of the bounding box surrounding the pedestrians.

Based on the findings, YOLOv7 demonstrates accurate human pose estimation in challenging conditions such as poor lighting, indoor and outdoor environments, occlusion, and partially cropped images. Nevertheless, it exhibits superior performance in outdoor and well-lit scenarios and when subjects are relatively closer to the camera. Moreover, this study introduces a novel pose evaluation metric, Mean Scaled Joint Error (MSJE), designed to address the distance bias of the traditional MPJPE metric. The results indicate that MSJE offers a more realistic evaluation of the model's performance.

**FUTURE DIRECTION AND NEXT STEPS**

In this study, we aimed to collect the existing datasets that captured human behavior in the presence of moving obstacles or robots equipped with data collection devices. However, these two datasets were originally collected for different purposes and were repurposed for human-robot interaction and pose estimation analysis. Besides, despite YOLOv7 showing relatively satisfactory performance, the results and error metrics indicate the presence of errors in the estimation.

To bridge the identified gap, the next step is collecting an enriched dataset through two sources: a field experiment and a controlled virtual immersive reality environment (VIRE) (*29*). The field experiment entails navigating our robot in shared urban and campus areas. This approach allows us to gather an in-the-wild dataset reflecting daily human activities while exposing the robot to specific features and scenarios essential to be captured for future socially aware path-planning purposes.

On the other hand, the VIRE experiment can offer precise joint location data by attaching sensors to the human body, granting full control over the experiment. Combining these data sources will result in an ideal dataset for conducting further research on pose prediction, pedestrian behavior analysis, and socially aware path planning for automated delivery robots.

Regarding methodology, investigation of the latest version of YOLO, version 8, and comparing the result with YOLOv7 is recommended. In future, we plan to release all the labeled data as an open-access repository.

**ACKNOWLEDGEMENT**

This study is supported by the NSERC Alliance grant and FuseForward.

# REFERENCES

1. Huai, J., Y. Qin, F. Pang, and Z. Chen, Segway DRIVE benchmark: Place recognition and SLAM data collected by a fleet of delivery robots. *arXiv preprint arXiv:1907.03424*, 2019.

2. Cregger, J., E. Machek, A. Epstein, T. Lennertz, J. Shaw, K. Dopart, M. Behan, et al., Emerging Automated Urban Freight Delivery Concepts: State of the Practice Scan, 2020.

3. Jennings, D. and M. Figliozzi, Study of sidewalk autonomous delivery robots and their potential impacts on freight efficiency and travel. *Transportation Research Record*, Vol. 2673, No. 6, 2019, pp. 317–326.

4. *INFO BRIEF Sharing Spaces with Robots: The Basics of Personal Delivery Devices.* `http://www.pedbikeinfo.org/cms/downloads/PBIC_InfoBrief_SharingSpaceswithRobots.pdf`, ????, accessed: 2022-08-12.

5. Wang, J., S. Tan, X. Zhen, S. Xu, F. Zheng, Z. He, and L. Shao, Deep 3D human pose estimation: A review. *Computer Vision and Image Understanding*, Vol. 210, 2021, p. 103225.

6. Fang, Z. and A. M. López, Intention recognition of pedestrians and cyclists by 2d pose estimation. *IEEE Transactions on Intelligent Transportation Systems*, Vol. 21, No. 11, 2019, pp. 4773–4783.

7. Marginean, A., R. Brehar, and M. Negru, Understanding pedestrian behaviour with pose estimation and recurrent networks. In *2019 6th International Symposium on Electrical and Electronics Engineering (ISEEE)*, IEEE, 2019, pp. 1–6.

8. Zheng, J., X. Shi, A. Gorban, J. Mao, Y. Song, C. R. Qi, T. Liu, V. Chari, A. Cornman, Y. Zhou, et al., Multi-modal 3d human pose estimation with 2d weak supervision in autonomous driving. In *Proceedings of the IEEE/CVF Conference on Computer Vision and Pattern Recognition*, 2022, pp. 4478–4487.

9. Bauer, P., A. Bouazizi, U. Kressel, and F. B. Flohr, Weakly Supervised Multi-Modal 3D Human Body Pose Estimation for Autonomous Driving. In *2023 IEEE Intelligent Vehicles Symposium (IV)*, IEEE, 2023, pp. 1–7.

10. Kalatian, A. and B. Farooq, A context-aware pedestrian trajectory prediction framework for automated vehicles. *Transportation research part C: emerging technologies*, Vol. 134, 2022, p. 103453.

11. Sárándi, I., A. Hermans, and B. Leibe, Learning 3D human pose estimation from dozens of datasets using a geometry-aware autoencoder to bridge between skeleton formats. In *Proceedings of the IEEE/CVF Winter Conference on Applications of Computer Vision*, 2023, pp. 2956–2966.

12. Carlevaris-Bianco, N., A. K. Ushani, and R. M. Eustice, University of Michigan North Campus long-term vision and lidar dataset. *The International Journal of Robotics Research*, Vol. 35, No. 9, 2016, pp. 1023–1035.

13. Milan, A., L. Leal-Taixé, I. Reid, S. Roth, and K. Schindler, MOT16: A benchmark for multi-object tracking. *arXiv preprint arXiv:1603.00831*, 2016.

14. Ess, A., B. Leibe, K. Schindler, and L. Van Gool, Robust multiperson tracking from a mobile platform. *IEEE transactions on pattern analysis and machine intelligence*, Vol. 31, No. 10, 2009, pp. 1831–1846.

15. Ess, A., B. Leibe, and L. Van Gool, Depth and appearance for mobile scene analysis. In *2007 IEEE 11th international conference on computer vision*, IEEE, 2007, pp. 1–8.

16.    Ess, A., B. Leibe, K. Schindler, and L. Van Gool, Moving obstacle detection in highly dynamic scenes. In *2009 IEEE International Conference on Robotics and Automation*, IEEE, 2009, pp. 56–63.

17.    Ess, A., B. Leibe, K. Schindler, and L. Van Gool, A mobile vision system for robust multi-person tracking. In *2008 IEEE Conference on Computer Vision and Pattern Recognition*, IEEE, 2008, pp. 1–8.

18.    Wang, C.-Y., A. Bochkovskiy, and H.-Y. M. Liao, YOLOv7: Trainable bag-of-freebies sets new state-of-the-art for real-time object detectors. In *Proceedings of the IEEE/CVF Conference on Computer Vision and Pattern Recognition*, 2023, pp. 7464–7475.

19.    Lugaresi, C., J. Tang, H. Nash, C. McClanahan, E. Uboweja, M. Hays, F. Zhang, C.-L. Chang, M. G. Yong, J. Lee, et al., Mediapipe: A framework for building perception pipelines. *arXiv preprint arXiv:1906.08172*, 2019.

20.    Bierlaire, M., MoveNet: Ultra fast and accurate pose detection model., ????

21.    Cao, Z., T. Simon, S.-E. Wei, and Y. Sheikh, Realtime multi-person 2d pose estimation using part affinity fields. In *Proceedings of the IEEE conference on computer vision and pattern recognition*, 2017, pp. 7291–7299.

22.    Vishnu, J. and S. Divya, A Comparative Study of Human Pose Estimation, ????

23.    Ionescu, C., D. Papava, V. Olaru, and C. Sminchisescu, Human3. 6m: Large scale datasets and predictive methods for 3d human sensing in natural environments. *IEEE transactions on pattern analysis and machine intelligence*, Vol. 36, No. 7, 2013, pp. 1325–1339.

24.    Katircioglu, I., C. Georgantas, M. Salzmann, and P. Fua, Dyadic human motion prediction. *arXiv preprint arXiv:2112.00396*, 2021.

25.    Wang, J., H. Xu, M. Narasimhan, and X. Wang, Multi-person 3d motion prediction with multi-range transformers. *Advances in Neural Information Processing Systems*, Vol. 34, 2021, pp. 6036–6049.

26.    Diller, C., T. Funkhouser, and A. Dai, Forecasting characteristic 3D poses of human actions. In *Proceedings of the IEEE/CVF Conference on Computer Vision and Pattern Recognition*, 2022, pp. 15914–15923.

27.    Cao, Z., H. Gao, K. Mangalam, Q.-Z. Cai, M. Vo, and J. Malik, Long-term human motion prediction with scene context. In *Computer Vision–ECCV 2020: 16th European Conference, Glasgow, UK, August 23–28, 2020, Proceedings, Part I 16*, Springer, 2020, pp. 387–404.

28.    Adeli, V., M. Ehsanpour, I. Reid, J. C. Niebles, S. Savarese, E. Adeli, and H. Rezatofighi, Tripod: Human trajectory and pose dynamics forecasting in the wild. In *Proceedings of the IEEE/CVF International Conference on Computer Vision*, 2021, pp. 13390–13400.

29.    Farooq, B., E. Cherchi, and A. Sobhani, Virtual immersive reality for stated preference travel behavior experiments: A case study of autonomous vehicles on urban roads. *Transportation research record*, Vol. 2672, No. 50, 2018, pp. 35–45.